%% file: paper.tex
\documentclass{article}
\usepackage{rotating}
\input{newcommands2}

\input{newcommands3}

\usepackage{dblfloatfix}    
\usepackage{graphicx}
\usepackage{comment}
\usepackage{amsmath,amssymb} 
\usepackage{color}

\usepackage{slashbox}

\usepackage{times}
\usepackage{epsfig}
\usepackage{amssymb}
\usepackage{balance}

\usepackage{units}
\usepackage{mathtools}
\usepackage[linesnumbered,lined,boxed]{algorithm2e}
\usepackage{upgreek}
\usepackage{colortbl}
\usepackage{stmaryrd}
\usepackage[mathscr]{euscript}

\definecolor{morado}{cmyk}{0,1,0.50,0}
\definecolor{gris2}{cmyk}{0,0,0,0.25}
\definecolor{gris}{cmyk}{0,0,0,0.1}
\definecolor{amarillo}{cmyk}{0,0,0.6,0}
\definecolor{blanco}{cmyk}{0,0,0,0}
\definecolor{negro}{cmyk}{1,1,1,0}
\definecolor{orange}{cmyk}{0,0.5,0.8,0}

\newcommand{\tscr}[1]{\boldsymbol{\mathscr{#1}}}
\newcommand{\nmat}[1]{\prescript{(n)}{}{\boldsymbol{#1}}} 
\usepackage{spconf,amsmath,graphicx}

\title{Multi-Dimensional Signal Recovery using Low-rank Deconvolution}
\name{David Reixach}
\address{Universitat Politècnica de Catalunya, BarcelonaTech, Spain}
\begin{document}
%
\maketitle
\begin{abstract}

\noindent In this work we present Low-rank Deconvolution, a powerful framework for low-level feature-map learning for efficient signal representation with application to signal recovery. Its formulation in multi-linear algebra inherits properties from convolutional sparse coding and low-rank approximation methods as in this setting signals are decomposed in a set of filters convolved with a set of low-rank tensors. We show its advantages by learning compressed video representations and solving image in-painting problems.

\end{abstract}
\begin{keywords}
Tensors, Sparse Coding, Low-rank Approximation, Tensor completion
\end{keywords}
\section{Introduction}
\label{sec:intro}

Convolutional Sparse Coding (CSC) aims to decompose an input signal $\bS$ as a sum of few learned features (dictionary) $\{\bD_m\}$ convolved with a set of sparse activation maps $\{\bX_m\}$ such that $ \bS \approx \sum_m \bD_m \ast \bX_m$. It achieved certain relevance once it was proved that it could be efficiently formulated in the frequency domain in terms of the Convolutional Basis Pursuit Denoising (CBPDN)~\cite{zeiler2010deconvolutional} as an optimization problem. As a result, this method has proven application in many areas such image denoising, in-painting, super-resolution among others~\cite{kavukcuoglu2010learning,zeiler2011adaptive,heide2015fast,gu2015convolutional,papyan2017convolutional}. Another technique commonly used to learn low/mid level features is Low-rank (LR) approximation, which relays on the fact that data obtained from many natural processes presents a low-rank structure. These methods are popular
for denoising and completion~\cite{ji2010robust,candes2012exact,candes2010matrix}. And also in the multi-linear algebra domain as low-rank tensor applications~\cite{long2019low,zhang2019nonlocal,cai2019nonconvex}.

Our approach considers a multi-linear convolutional model with sparse and low-rank activation maps that inherits properties from both CSC and LR. The idea is not new as we follow~\cite{humbert2020lowrank,humbert2021tensor}. However, our contribution goes further as we present a powerful linear-algebra formulation that allows for learning complex low-rank activations, for a given set of features, which permit us to represent multidimensional signals more efficiently than classical CSC approaches, also with direct application in tensor completion settings which we prove by learning compressed representations of video sequences and solving image in-painting problems. And finally we show that in this framework the sparsity constraint can be avoided allowing for simple algorithms to solve the optimization problem.



\subsection{Notation}
\label{sec:note}

Let $\tscr{K} \in \mathbb{R}^{I_1\times I_2 \times\ldots\times I_N}$ be a $N$-order tensor. The PARAFAC~\cite{harshman1970foundations} decomposition (a.k.a. CANDECOMP~\cite{carroll1970analysis}) is defined as:
\begin{equation}\label{eq:parafac}
	\tscr{K} \approx \sum_{r=1}^{R}\mu_r  \bv_r^{(1)} \circ \bv_r^{(2)} \circ \ldots \circ \bv_r^{(N)},
\end{equation}
where $\bv_r^{(n)} \in \mathbb{R}^{I_n} $ with $n=\{1,\ldots,N\}$ and $\mu_r \in \mathbb{R}$ with $r=\{1,\ldots,R\}$, represent an one-order tensor and a weight coefficient, respectively. $\circ$ denotes an outer product of vectors. Basically, Eq.~\eqref{eq:parafac} is a rank-$R$ decomposition of $\tscr{K}$ by means of a sum of $R$ rank-$1$ tensors. If we group all these vectors per mode $(n)$, as $\bX^{(n)} = \big[ \bv_1^{(n)}, \bv_2^{(n)}, \ldots, \bv_R^{(n)} \big]$, we can define the Kruskal operator~\cite{kolda2006multilinear} as follows: 
\begin{equation}\label{eq:kruskal}
\llbracket \bX^{(1)}, \bX^{(2)}, \ldots, \bX^{(N)} \rrbracket =  \sum_{r=1}^{R} \bv_r^{(1)} \circ \bv_r^{(2)} \circ \ldots \circ \bv_r^{(N)},
\end{equation}
being the same expression as Eq.~\eqref{eq:parafac} with $\mu_r=1$ for $\forall r$, i.e., depicting a rank-$R$ decomposable tensor.

For later computations, we also define a matricization transformation to express tensors in a matrix form. Particularly, we will use an special case of matricization known as $n$-mode matricization~\cite{bader2006algorithm,kolda2006multilinear}. To this end, let $\mathcal{C} = \{c_1,\ldots,c_G\} =\{1,\dots,n-1,n+1,\dots,N\}$ be the collection of ordered modes different than $n$, and $\Lambda = \nicefrac{\prod_t I_t}{I_n}$ be the product of its correspondent dimensions; we can express then tensor $\tscr{K}$ in a matricized array as $\prescript{(n)}{}{\bK} \in \mathbb{R}^{I_n\times\Lambda}$. Note that we represent the $n$-mode matricization by means of a left super-index. The $n$-mode matricization is a mapping from the indices of $\tscr{K}$ to those of $\prescript{(n)}{}{\bK}$, defined as:
\begin{equation}\label{tensorm1}
	\big(\prescript{(n)}{}{\bK}\big)_{i_n,j} = \tscr{K}_{i_1,i_2,\dots,i_N} \, ,
\end{equation}
with:
\begin{equation}\label{tensorm2}
	j = 1+\sum_{g=1}^{G}\big[(i_{c_g}-1)\prod_{g'=1}^{G-1}I_{c_{g'}} \big].
\end{equation}

With these ingredients, and defining $\tscr{J}(\bX^{(1)}, \dots, \bX^{(N)}) = \llbracket \bX^{(1)}, \dots, \bX^{(N)} \rrbracket$, we can obtain the $n$-mode matricization of the Kruskal operator as: 
\begin{equation}\label{eq:kruskal_m}
\prescript{(n)}{}{\bJ} = \bX^{(n)}(\bQ^{(n)})^\top, 
\end{equation}
with:
\begin{equation}\label{eq:Q}
\bQ^{(n)} = \bX^{(N)}\odot \ldots \odot  \bX^{(n+1)}\odot  \bX^{(n-1)}\odot \ldots \odot   \bX^{(1)},
\end{equation}
where $\odot$ denotes the Khatri-Rao product~\cite{kolda2006multilinear}.

Finally, we can express the vectorized version of Eq.~\eqref{eq:kruskal_m} as: 
\begin{equation}
\label{eq:kruskal_v}
\mathrm{vec}\left(\prescript{(n)}{}{\bJ}\right) =
\big[\bQ^{(n)}\otimes \bI_{I_n}\big]\mathrm{vec}(\bX^{(n)}),
\end{equation}
where $\otimes$ indicates the Kronecker product, and $\mathrm{vec}(\cdot)$ is a vectorization operator. It is worth noting that doing so, the vectorized form of the Kruskal operator is represented by a linear expression.

\section{Low-rank Deconvolution}
\label{sec:lrd}

We now derive the formulation of our approach. Let $\tscr{S} \in \mathbb{R}^{I_1\times I_2\times\cdots\times I_N}$ be a multidimensional signal. Our goal is to obtain a multidimensional convolutional representation $\tscr{S}\approx\sum_m \tscr{D}_m \ast \tscr{K}_m$, where  $\tscr{D}_m \in \mathbb{R}^{L_1\times L_2\times\cdots\times L_N}$ acts as a dictionary, and $\tscr{K}_m \in \mathbb{R}^{I_1\times I_2\times\cdots\times I_N}$, the activation map, is a low-rank factored tensor (\ie a Kruskal tensor). If we write $ \tscr{K}_m =  \llbracket \bX_m^{(1)}, \bX_m^{(2)}, \ldots, \bX_m^{(N)} \rrbracket $ with $  \mathbf{X}_m^{(n)} \in \mathbb{R}^{I_n\times R} $, in the context of CBPDN we can obtain a non-convex problem as:
\begin{align}\label{eq:main}
	\argmin_{\{\bX_m^{(n)}\}}
	\frac{1}{2}&\left\lVert \sum_{m = 1}^{M}	\tscr{D}_m \ast \llbracket \bX_m^{(1)},\ldots,\bX_m^{(N)} \rrbracket -\tscr{S} \right\rVert_2^2 \nonumber\\
	&+ \sum_{m = 1}^{M}\sum_{n = 1}^{N}\lambda \left\lVert \bX_m^{(n)} \right\rVert_1,
\end{align}
where $\ast$ indicates a $N$-dimensional convolution. Following standard strategies to resolve the PARAFAC decomposition~\cite{kolda2006multilinear}, we propose to solve for each Kruskal factor $(n)$ alternately. We denoted this strategy as Low-rank Deconvolution (LRD).


 


\subsection{ADMM Algorithm}
\label{sec:admm}

In order to make
the optimization problem in Eq.~\eqref{eq:main} tractable, we rewrite it in a form suitable for Alternating Direction Method of Multipliers (ADMM)~\cite{boyd2011distributed} by using a couple of auxiliary variables $\{\bY_m^{(n)}\}$ and $\{\bU_m^{(n)}\}$ of the same size as $\{\bX_m^{(n)}\}$, obtaining the optimization as:
\begin{align}\label{eq:admm}
\argmin_{\{\bX_m^{(n)}\},\{\bY_m^{(n)}\}}
\frac{1}{2}&\left\lVert \sum_{m = 1}^{M}	\tscr{D}_m \ast \llbracket \ldots,\bX_m^{(n)},\ldots \rrbracket -\tscr{S} \right\rVert_2^2 \nonumber\\
&+ \sum_{m = 1}^{M}\lambda \left\lVert \bY_m^{(n)} \right\rVert_1\\
&\hspace{-1.8cm}\textrm{subject to} \hspace{0.5cm} \bX_m^{(n)} = \bY_m^{(n)} \,\,\forall \,m \nonumber
\end{align}

The previous problem can be carried out efficiently by solving each Kruskal factor $(n)$ alternately, updating one variable while fixing the others. Algorithm~\ref{algorithm_ADMM_based} explains the details. To initialize the auxiliary variables $\{\bY_m^{(n)}\}$ and $\{\bU_m^{(n)}\}$ and the selection of the penalty coefficient $\rho$, we adopt the proposal in~\cite{wohlberg2017sporco}, with an adaptive strategy for the latter.

\IncMargin{1em}
\begin{algorithm}[t!]
\begin{small}
\While{not converged}{
$\bX_m^{(n)(k+1)} = 
\argmin
\frac{1}{2}\left\lVert \sum_{m = 1}^{M}	\tscr{D}_m \ast \llbracket \ldots,\bX_m^{(n)},\ldots \rrbracket -\tscr{S} \right\rVert_2^2 
\plusc\frac{\rho}{2}\sum_{m = 1}^{M}\left\lVert \bX_m^{(n)} -\bY_m^{(n)(k)} + \bU_m^{(n)(k)} \right\rVert_2^2$

$\bY_m^{(n)(k+1)} = \mathbf{prox_1}_{\tfrac{\lambda}{\rho}}(\bX_m^{(n)(k+1)}+ \bU_m^{(k)})$

$\bU_m^{(n)(k+1)} = \bU_m^{(n)(k)} + \bX_m^{(n)(k+1)}-\bY_m^{(n)(k+1)}$

}
Not.: $\mathbf{prox_1}_\gamma(\bu) = \mathrm{sign}(\bu)\oplus \mathrm{max}(0,\lvert \bu\rvert-\gamma )$. 
\end{small}
\caption{\textbf{ADMM algorithm for LRD} for solving Eq.~\eqref{eq:admm}, considering a $(n)$-mode sub-problem. $\mathbf{prox}$ is proximal operator to perform a shrinkage. $\mathrm{sign(\cdot)}$, $\mathrm{max(\cdot)}$ and $\lvert \cdot \rvert$ of a vector considered to be applied element-wise. $\oplus$ denotes the element-wise product. Step 2 is solved applying section~\ref{section_for_DFT_domain}.}\label{algorithm_ADMM_based}
\end{algorithm}\DecMargin{1em}



\subsection{Formulation in the DFT Domain}\label{section_for_DFT_domain}

As it was discussed in~\ref{sec:intro}, the common practice to address CBPDN is to solve it in a DFT domain, achieving a solution both efficient and accurate. To this end, we denote by $\hat{\bA}$ an arbitrary variable $\bA$ in the DFT domain. Looking for a linear expression, let $\hat{{\bD}}^{(n)}_m =\mathrm{diag}\big(\mathrm{vec}\big(\nmat{\hat{D}}_m\big)\big) \in \mathbb{R}^{\Lambda I_n\times\Lambda I_n}$ be a linear operator for computing convolution, and $\hat{\bx}_m^{(n)} = \mathrm{vec}(\hat{\bX}_m^{(n)})\in \mathbb{R}^{MRI_n}$ be the vectorized Kruskal factor. We can define now $ \hat{\mathbf{Q}}_m^{(n)} =  \hat{\mathbf{X}}_m^{(N)}\odot \cdots \odot  \hat{\mathbf{X}}_m^{(n+1)}\odot  \hat{\mathbf{X}}_m^{(n-1)}\odot \cdots \odot \hat{\mathbf{X}}_m^{(1)} \in \mathbb{R}^{\Lambda\times R} $, as it was done in Eq.~\eqref{eq:Q}, with $\Lambda$ defined in section~\ref{sec:note}.

\begin{figure*}[!b]
\centering
\resizebox{17.4cm}{!} {
\begin{tabular}{@{}cccccccccc@{}}
 \includegraphics[clip, angle=0]{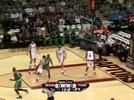}&
 \includegraphics[clip, angle=0]{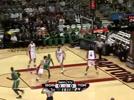}&
 \includegraphics[clip, angle=0]{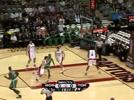}&
 \includegraphics[clip, angle=0]{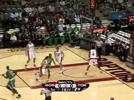}&
  \includegraphics[clip, angle=0]{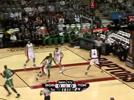}&
 \includegraphics[clip, angle=0]{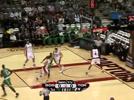}&
 \includegraphics[clip, angle=0]{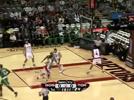}&
 \includegraphics[clip, angle=0]{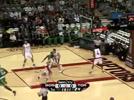}&
  \includegraphics[clip, angle=0]{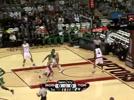}&
 \includegraphics[clip, angle=0]{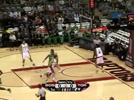}\\
  \includegraphics[clip, angle=0]{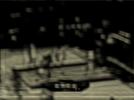}&
 \includegraphics[clip, angle=0]{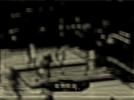}&
 \includegraphics[clip, angle=0]{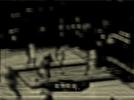}&
 \includegraphics[clip, angle=0]{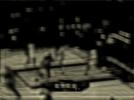}&
  \includegraphics[clip, angle=0]{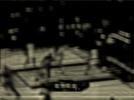}&
 \includegraphics[clip, angle=0]{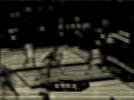}&
 \includegraphics[clip, angle=0]{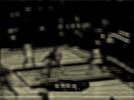}&
 \includegraphics[clip, angle=0]{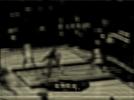}&
  \includegraphics[clip, angle=0]{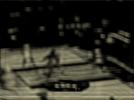}&
 \includegraphics[clip, angle=0]{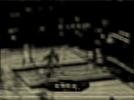}\\
   \includegraphics[clip, angle=0]{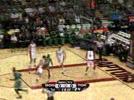}&
 \includegraphics[clip, angle=0]{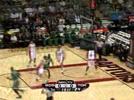}&
 \includegraphics[clip, angle=0]{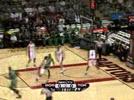}&
 \includegraphics[clip, angle=0]{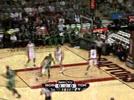}&
  \includegraphics[clip, angle=0]{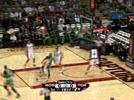}&
 \includegraphics[clip, angle=0]{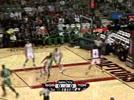}&
 \includegraphics[clip, angle=0]{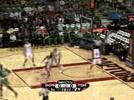}&
 \includegraphics[clip, angle=0]{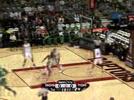}&
  \includegraphics[clip, angle=0]{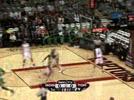}&
 \includegraphics[clip, angle=0]{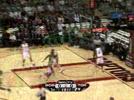}\\
\end{tabular}}
\vspace{-0.1cm}
  \caption{\textbf{Qualitative evaluation on RGB Basketball video.} In all cases, we show ten consecutive video frames. \textbf{Top.} Ground truth color frames. \textbf{Middle.} Color Video reconstruction using~\cite{wohlberg2017sporco}. \textbf{Bottom.} Our solution. As it can be seen, our method provides a more visually correct solution than~\cite{wohlberg2017sporco} that includes a wide variety of artifacts. Best viewed in color.}
\label{fig:reconstruction}
\end{figure*}


Assuming that boundary effects are negligible, i.e., relying on the use of filters of small spatial support $L_n$ for $n = 1,\dots,N$, we can formulate the problem in Alg.~\ref{algorithm_ADMM_based}-step 2 in the DFT domain as:
\begin{align}\label{eq:admm_dft}
	\argmin_{\{\hat{\bx}_m^{(n)}\}}
	\frac{1}{2}&\left\lVert \sum_{m = 1}^{M}	\hat{\bD}_m^{(n)} \big[ \hat{\bQ}_m^{(n)}\otimes \bI_{I_n}\big] \hat{\bx}_m^{(n)} -\hat{\bs}^{(n)} \right\rVert_2^2 \nonumber\\
	&+ \frac{\rho}{2}\sum_{m = 1}^{M}\left\lVert \hat{\bx}_m^{(n)} -\hat{\bz}_m^{(n)} \right\rVert_2^2 ,
\end{align}
where we have used a shortcut variable $ \hat{\mathbf{z}}_m^{(n)} = \hat{\mathbf{y}}_m^{(n)} - \hat{\mathbf{u}}_m^{(n)} $ that encompasses vectorized versions of $ \hat{\mathbf{Y}}_m^{(n)}$ and $ \hat{\mathbf{U}}_m^{(n)}$, respectively. Moreover, $\hat{\mathbf{s}}^{(n)}$ is the vectorized version of $\prescript{(n)}{}{\hat{\bS}}$. To solve the problem, we first define some supporting expressions:
\begin{align}
	\hat{\bW}_m^{(n)} &= \hat{\bD}_m^{(n)}\big[\hat{\bQ}_m^{(n)}\otimes \bI_{I_n}\big] \label{eq:dft_wm} ,\\
	\hat{\bW}^{(n)} &= \big[\hat{\bW}_0^{(n)},\hat{\bW}_1^{(n)},\ldots,\hat{\bW}_M^{(n)}\big] \label{eq:dft_w} ,\\
	\hat{\bx}^{(n)} &= \big[(\hat{\bx}_0^{(n)})^\top,(\hat{\bx}_1^{(n)})^\top,\ldots,(\hat{\bx}_M^{(n)})^\top\big]^\top \label{eq:dft_x} ,\\
	\hat{\bz}^{(n)}& = \big[(\hat{\bz}_0^{(n)})^\top,(\hat{\bz}_1^{(n)})^\top,\ldots,(\hat{\bz}_M^{(n)})^\top\big]^\top \label{eq:dft_z},
\end{align}
in order to finally transform the problem in Eq.~\eqref{eq:admm_dft} into:
\begin{equation}\label{eq:admm_dft2}
	\hspace{-0.2cm}\argmin_{\hat{\bx}^{(n)}}
	\frac{1}{2}\left\lVert	\hat{\bW}^{(n)} \hat{\bx}^{(n)} -\hat{\bs}^{(n)} \right\rVert_2^2
	+ \frac{\rho}{2}\left\lVert \hat{\bx}^{(n)} -\hat{\bz}^{(n)} \right\rVert_2^2 .
\end{equation}

Fortunately, Eq.~\eqref{eq:admm_dft2} can be solved in closed form, by means of the next linear system:
\begin{equation}\label{eq:admm_sol}
	\hspace{-0.2cm}\big[(\hat{\bW}^{(n)})^H \hat{\bW}^{(n)} +\rho \bI_{\beta}\big]
	\hat{\bx}^{(n)} = (\hat{\bW}^{(n)})^H \hat{\bs}^{(n)}
	+ \rho \hat{\bz}^{(n)},
\end{equation}
where $(\cdot)^H$ denotes a conjugate transpose matrix, and $\beta=MRI_n$. It is worth pointing out that the Kruskal tensor in Eq.~\eqref{eq:kruskal} can be seen as a separable filter, and hence allows for the DFT transform to be computed independently for each factor.



\IncMargin{1em}
\begin{algorithm}[t!]
\SetKwInOut{Input}{input}\SetKwInOut{Output}{output}
\Input{$\tscr{S}$, $\{\tscr{D}_m\}_{m=1}^M,\{\bX_{0,m}^{(n)}\}_{n=1,m=1}^{N,M}, R>0$ }
\Output{ $\{\bX_m^{(n)}\}_{n=1,m=1}^{N,M}$ }

\tcc{Initialize Kruskal Factors}
$\{\bX_m^{(n)}\} = \{\bX_{0,m}^{(n)}\} $\\  
\tcc{Main Loop, Eq.~\eqref{eq:main}}
\While{not converged}{
\For{$n = 1,\dots,N$}{
{\scriptsize $\bX_m^{(n)}=\argmin
    \frac{1}{2}\left\lVert \sum_{m = 1}^{M}	\tscr{D}_m \ast \llbracket \bX_m^{(1)},\ldots,\bX_m^{(N)} \rrbracket -\tscr{S} \right\rVert_2^2  +
    \Phi(\{\bX_m^{(n)}\})$}
}
}
\caption{ \textbf{LRD algorithm}. Here $\Phi(\cdot)$ refers to the choice of regularization. If norm-1 is choosen, step 4 is solved by using Alg.~\ref{algorithm_ADMM_based}, else if norm-2 is choosen, step 4 is performed by directly solving the linear system in eq.~\eqref{eq:admm_l2_sol}. The full algorithm solves the LRD problem by means of an alternated approach for every $n$-mode.}\label{algorithm_MAIN}
\end{algorithm}\DecMargin{1em}

\subsubsection{Choice of Regularization}\label{section:regularization}

We suspect that in our case the sparsity regularization is not required as the low-rank constraint might be sufficient. One can observe that replacing the norm-$1$ term by a squared norm-$2$ term and $\lambda$ by $\nicefrac{\alpha}{2}$ in eq.~\eqref{eq:admm}, the solution is then given by:
\begin{equation}\label{eq:admm_l2_sol}
	\hspace{-0.2cm}\big[(\hat{\bW}^{(n)})^H \hat{\bW}^{(n)} +\alpha \bI_{\beta}\big]
	\hat{\bx}^{(n)} = (\hat{\bW}^{(n)})^H \hat{\bs}^{(n)},
\end{equation}

with $\beta$ defined in section~\ref{section_for_DFT_domain}. Then, the full LRD method is summarized in algorithm~\ref{algorithm_MAIN}. As it can be seen, we solve for every $n$-mode independently.

\begin{figure*}[t!]
\centering
\resizebox{17.4cm}{!} {
\begin{tabular}{@{}ccccc@{}}
 \includegraphics[viewport=40 200 540 640, clip, angle=0]{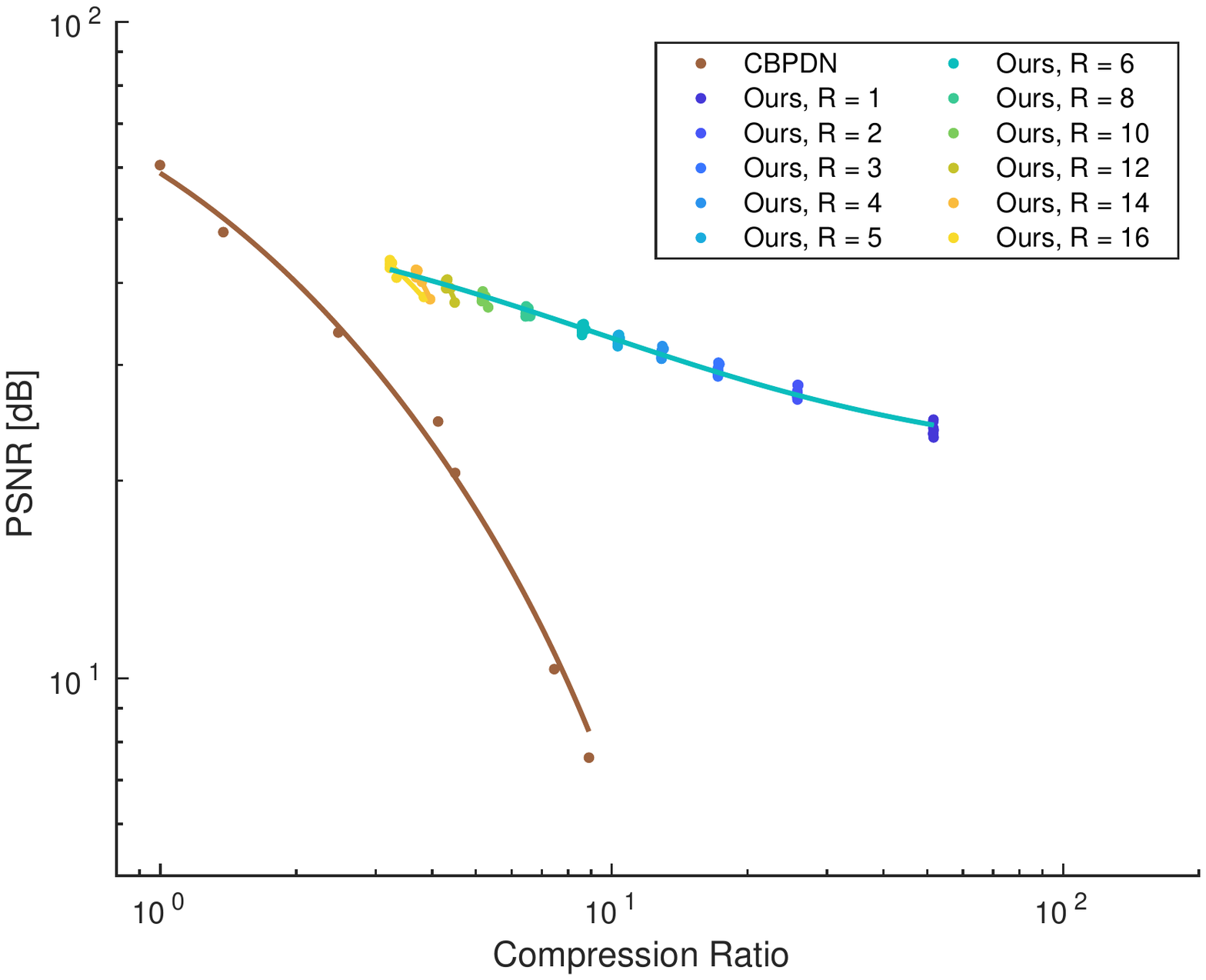}&
 \includegraphics[viewport=40 200 540 640, clip, angle=0]{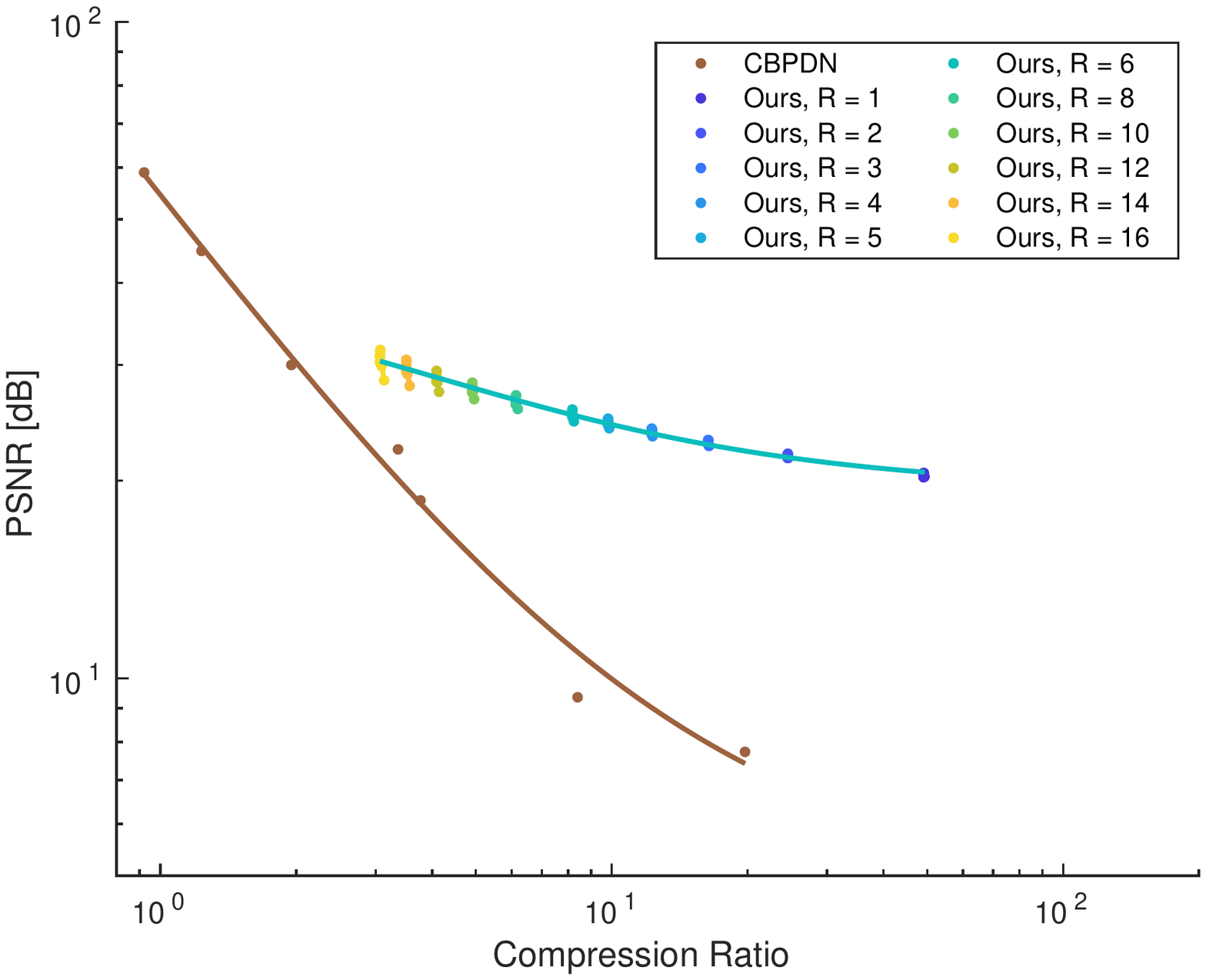}&
 \includegraphics[viewport=40 200 540 640, clip, angle=0]{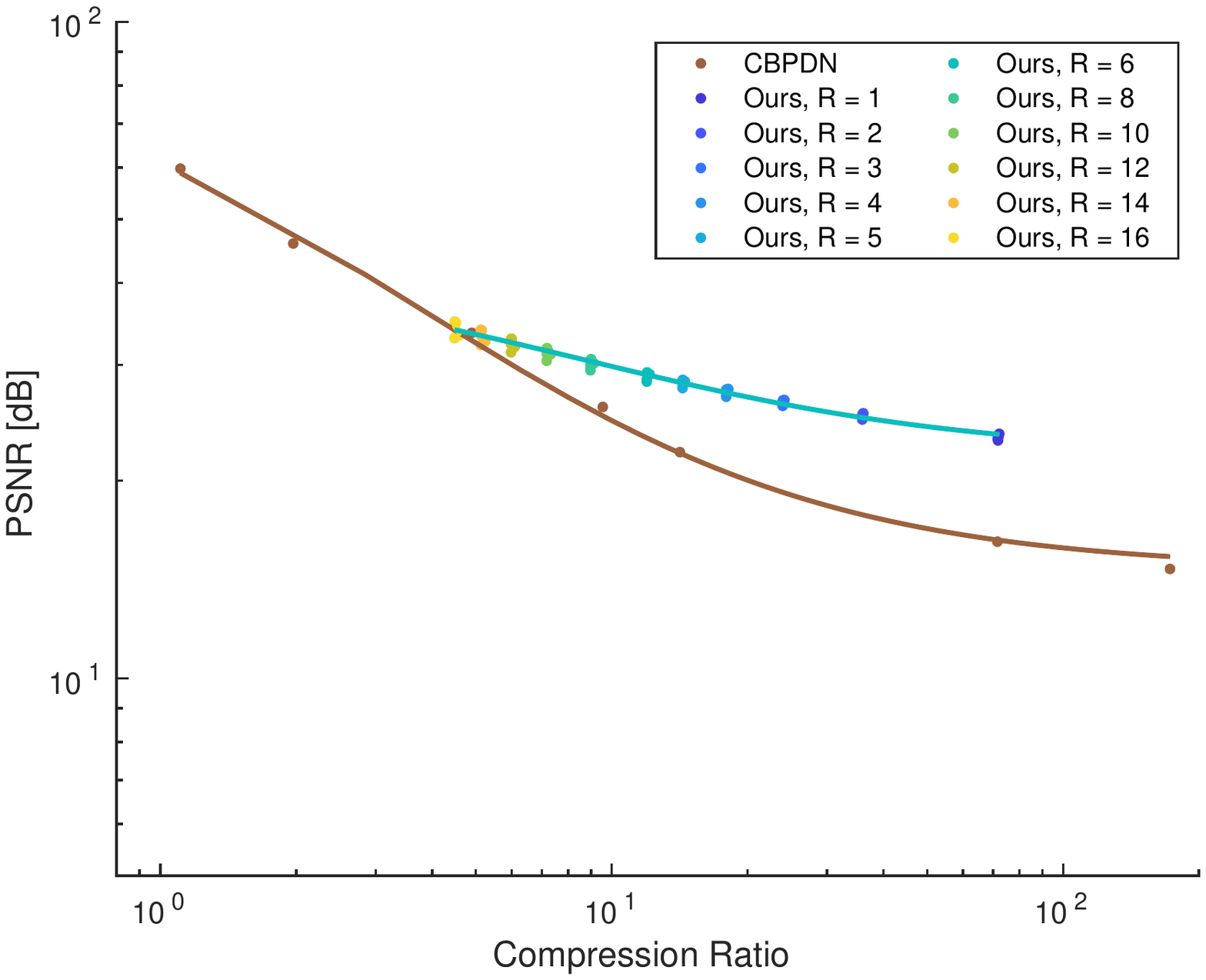}&
 \includegraphics[viewport=40 200 540 640, clip, angle=0]{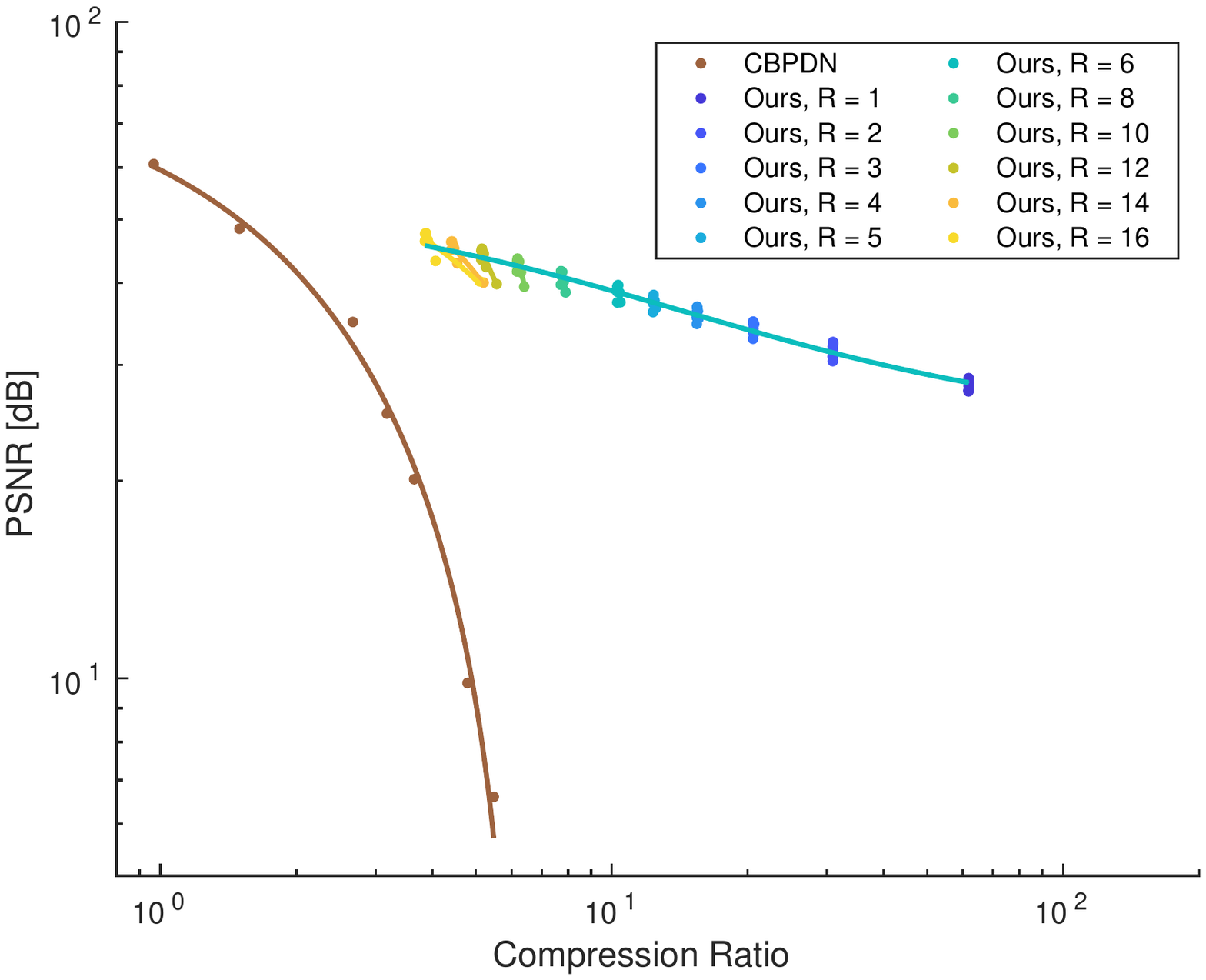}&
 \includegraphics[viewport=40 200 540 640, clip, angle=0]{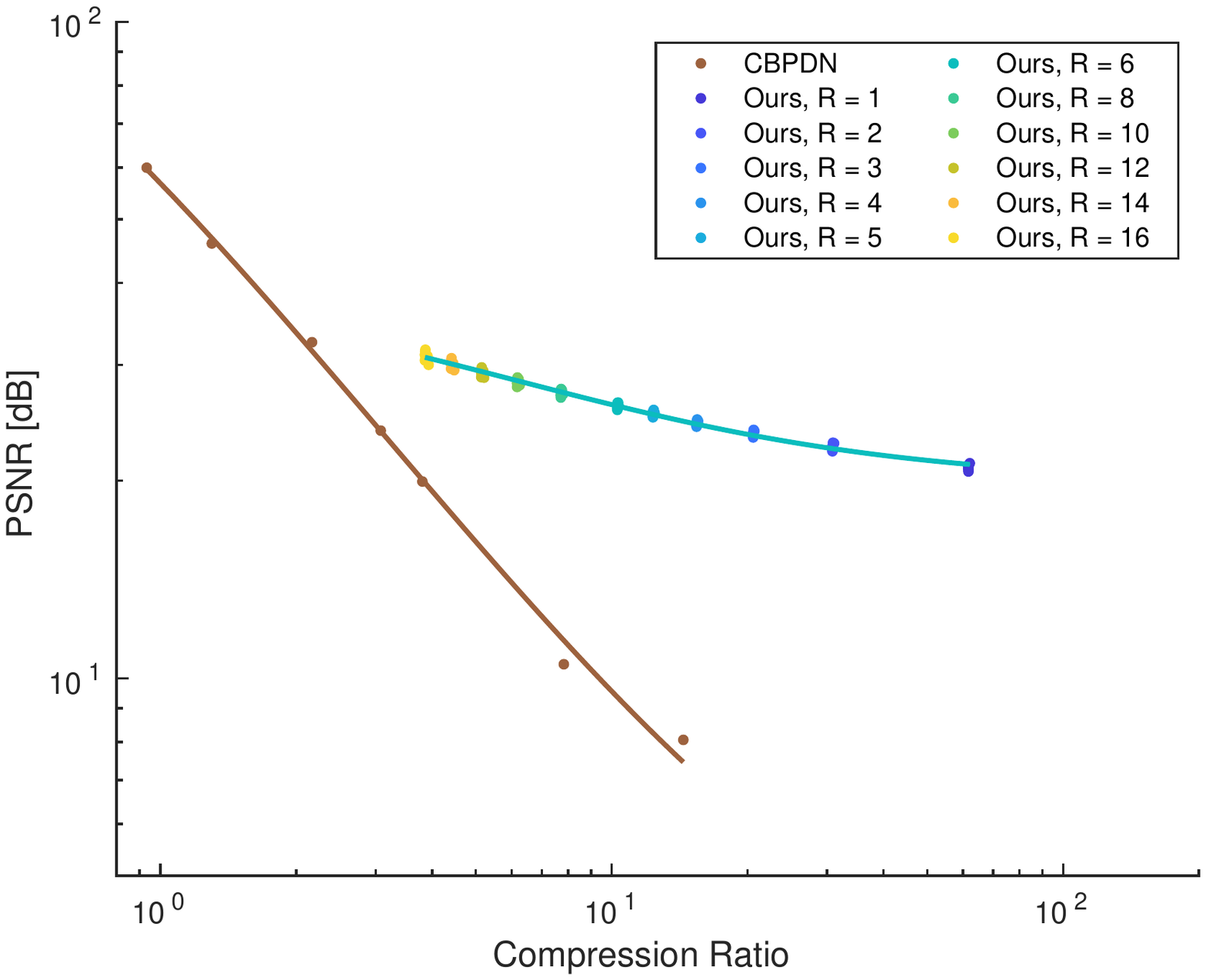}
\end{tabular}}
\vspace{-0.1cm}
\caption{\textbf{Quality of reconstruction (PSNR) vs. compression rate (CR).} We display results on the videos Basketball, Football1, Ironman, Skiing and Soccer, respectively. PSNR evolution as a function of CR for CBPDN~\cite{wohlberg2017sporco} and our approach for different $R$ values. Best viewed in color.}
\label{fig:otb_global}
\end{figure*}

\subsection{Linear Mask Decoupling for Tensor Completion}\label{section:completion}

We continue by applying LRD to tensor completion problems. To do so, we are required to formulate the optimization problem masking out the unknowns which are given in the spatial domain. This requires us to include the DFT transform in our formulation and to consider the spatial version of the signal $\bs^{(n)}$:

\begin{equation}\label{eq:maskDec}
	\hspace{-0.2cm}\argmin_{\hat{\bx}^{(n)}}
	\frac{1}{2}\left\lVert	\bP^{(n)}\hat{\bF}^{(n)} \hat{\bW}^{(n)} \hat{\bx}^{(n)} -\bs^{(n)} \right\rVert_2^2
	+ \frac{\alpha}{2}\left\lVert \hat{\bx}^{(n)}\right\rVert_2^2 .
\end{equation}

Here, $\hat{\bF}^{(n)} = \hat{\bF}_N\otimes\dots\otimes\hat{\bF}_{n+1}\otimes\hat{\bF}_{n-1}
\otimes\dots\otimes\hat{\bF}_1\otimes\hat{\bF}_n $ is the matricization of the multilinear DFT inverse transform being $\hat{\bF}_i$ the inverse transform for mode-$i$ and $\bP^{(n)}$ the mask matrix, which is diagonal for a tensor completion problem. Then, its solution is given by, 

\begin{equation}\label{eq:maskDec_sol}
	\hspace{-0.2cm}\big[(\hat{\bT}^{(n)})^H \hat{\bT}^{(n)} +\alpha \bI_{\beta}\big]
	\hat{\bx}^{(n)} = (\hat{\bT}^{(n)})^H \bs^{(n)},
\end{equation}

with $\hat{\bT}^{(n)} = \bP^{(n)}\hat{\bF}^{(n)}\hat{\bW}^{(n)}$ and $\beta$ defined in section~\ref{section_for_DFT_domain}. One can see that algorithm~\ref{algorithm_MAIN} can be easily extended to consider such approach.

\section{Experiments}
\label{sec:exp}


\subsection{Compressed Video Reconstruction}

We select five color sequences of the OTB50 dataset~\cite{wu2013online}, denoted as {\em Basketball}, {\em Football1}, {\em Ironman}, {\em Soccer} and {\em Skiing}. For every video, we consider the first 78 frames, with a minimum resolution of $[30\times 30]$ pixels, and we represent it by means of a $3$-order tensor with 3 channels. Dictionary dimensions are chosen as $\tscr{D}_{m,c} \in \mathbb{R}^{L_1\times L_2\times L_3}$ for $m = \{1,\dots,M\}$ and $c = \{1,\dots,C\}$, with $\{L_n = 5\}_{n=1}^3 $, $M=25$ and $C=3$. 
We split every sequence in two: we use the first half for learning the filters applying the algorithm from~\cite{garcia2018convolutional}, and the second half for testing.

To show the reconstruction quality we use a Peak Signal-to-Noise Ratio (PSNR), together with a Compression Ratio (CR) defined as $CR = \nicefrac{\prod_{n=1}^N I_n}{N_{NZ}}$
where $N_{NZ} =  \sum_m \left\lVert \bX_m \right\rVert_1$ for the CBPDN approach~\cite{wohlberg2017sporco} and $N_{NZ} =  \sum_m\sum_n \left\lVert \bX_m^{(n)} \right\rVert_1$ for our method. This measure is equivalent to the inverse of the number of features relative to the size of the signal which states the efficiency of the representation. Additionally, we also report results as a function of $\lambda$ for all methods, and evaluate for different rank values $R$ in our case. 


Figure~\ref{fig:otb_global} 
shows that our approach obtains a good reconstruction even for high levels of compression. 
While CBPDN~\cite{wohlberg2017sporco} obtains good results for small compression, quickly dropping as more sparsity is demanded (by increasing $\lambda$). 
Recall that the low-rank factorization is an important source of compression as according to section~\ref{sec:note} a rank-$R$ tensor is equivalent to the sum of $R$ rank-$1$ tensors, which in turn can be expressed as $\sum_nI_n$ instead of $\prod_nI_n$ values.
Finally, we present a qualitative comparison on the Basketball sequence in Fig.~\ref{fig:reconstruction} for a 
similar $CR\approx9$, the obtained accuracy is $PSNR=10.79$ and $PSNR=28.32$ for the CBPDN~\cite{wohlberg2017sporco} and our method, respectively. Again, we can observe how our estimation is visually more accurate than that provided by competing techniques.

\subsection{Image In-painting}

For this problem, we consider ten grey-scale images from~\cite{gu2014weighted} resized to $[100\times 100]$ pixels, named {\em Barbara}, {\em Boat}, {\em Cameraman}, {\em Couple}, {\em Fingerprint}, {\em Hill}, {\em House}, {\em Man}, {\em Montage} and {\em Peppers}. For each of them we mask out a random sample of the pixels with a proportion of $\{30\%,50\%,60\%\}$ relative to the total number of pixels of the image and we use the method from section~\ref{section:completion} to recover the original signal, with parameters set to $\{R=3,  M=15, \alpha = 10^{-4}\}$ and filters learned on the city and fruit datasets from~\cite{zeiler2010deconvolutional}. Results are presented in Table~\ref{table_inpainting}. Both the qualitative and quantitative results show that our method is capable of recovering the original signal even for an important number of missing entries.

\begin{table}[t!]
\centering
\resizebox{8cm}{!} {
\begin{tabular}{@{}ccc@{}}
 \includegraphics[viewport=200 320 420 580, clip, angle=0]{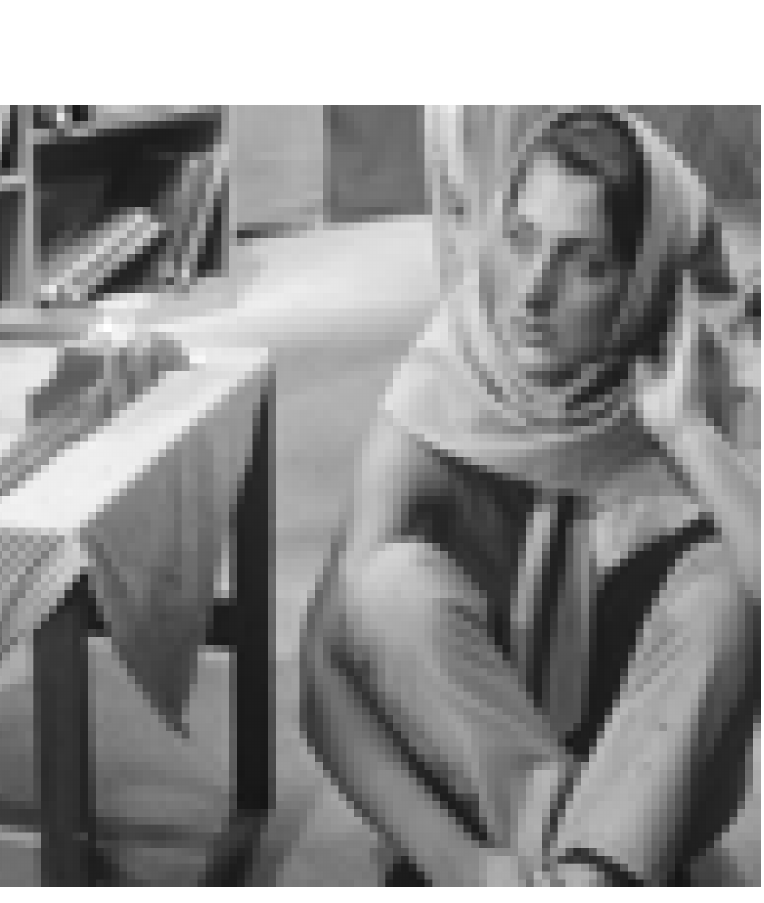}&
 \includegraphics[viewport=200 320 420 580, clip, angle=0]{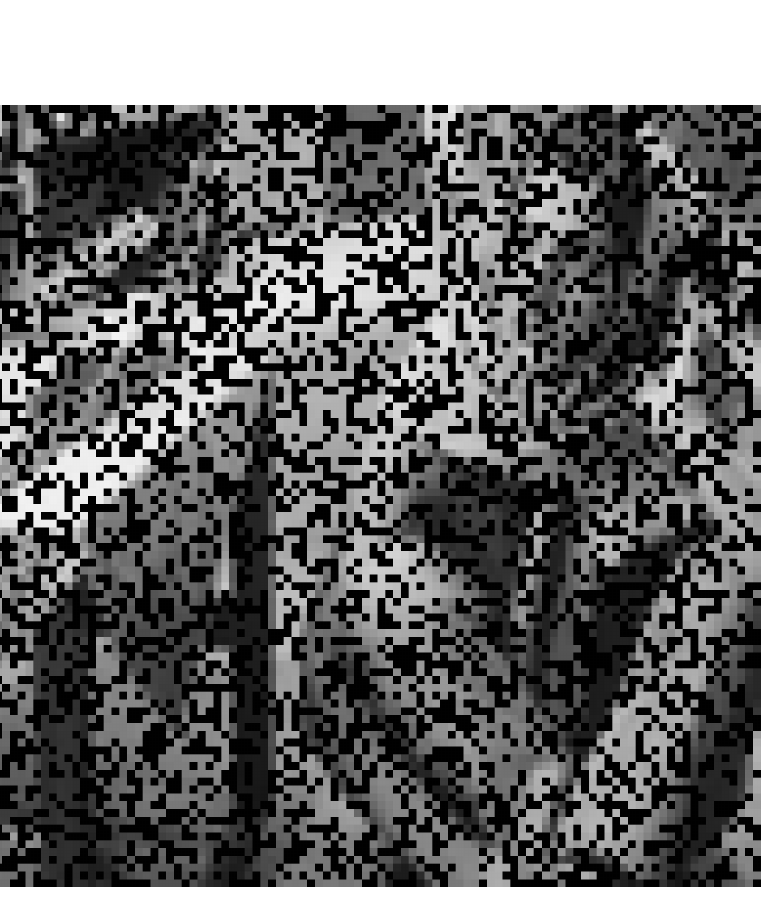}&
 \includegraphics[viewport=200 320 420 580, clip, angle=0]{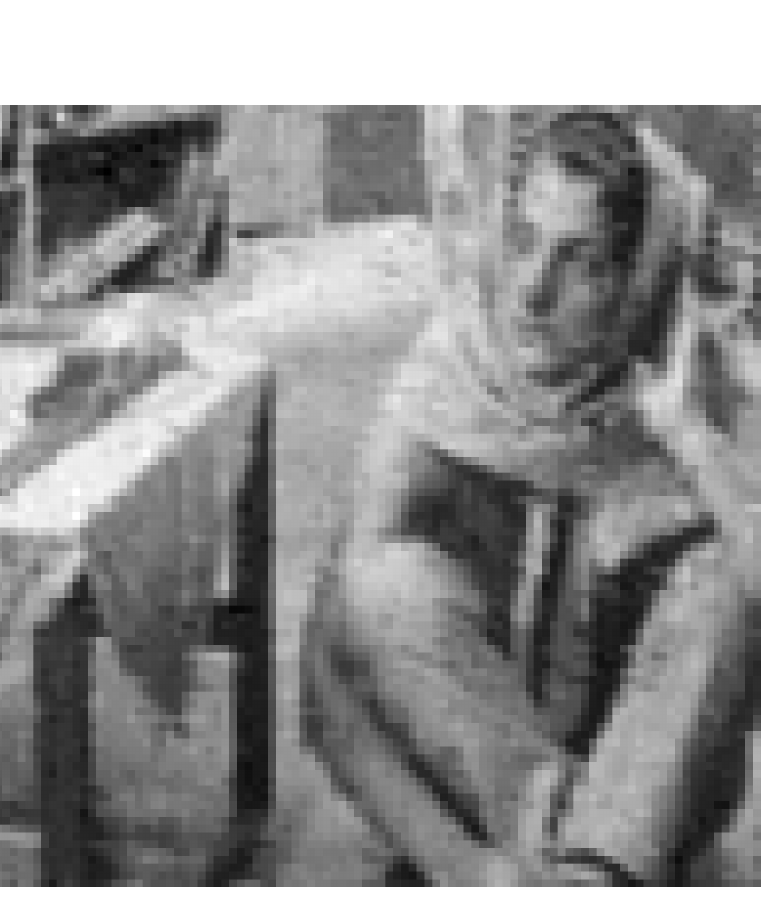}\\
\end{tabular}}\\
\centering
\resizebox{8cm}{!} {
\begin{tabular}{||c|| c | c | c | c | c | c | c | c | c | c | c ||}\hline
{\backslashbox{Missing}{Image}} & {\rotatebox[origin=c]{90}{Barbara}} & {\rotatebox[origin=c]{90}{Boat}} & {\rotatebox[origin=c]{90}{C.Man}} & {\rotatebox[origin=c]{90}{Couple}} & {\rotatebox[origin=c]{90}{F.Print}} & {\rotatebox[origin=c]{90}{Hill}} & {\rotatebox[origin=c]{90}{House}} & {\rotatebox[origin=c]{90}{Man}} & {\rotatebox[origin=c]{90}{Montage}} & {\rotatebox[origin=c]{90}{Peppers}} & {\cellcolor{gris2}\rotatebox[origin=c]{90}{Av.}}\\
\hline
\hline
{30\%} & {26.80} & {23.64} & {26.96}  & {24.29}  & {20.20}  & {25.86}  & {30.28}  & {22.17}  & {27.74}  & {23.13}  & \cellcolor{gris2}{25.11} \\
\hline
{50\%} & {23.76} & {23.10} & {24.83}  & {22.84}  & {18.17}  & {23.41}  & {27.32}  & {21.36}  & {23.33}  & {20.76}  & \cellcolor{gris2}{22.89} \\
\hline
{60\%} & {22.48} & {22.54} & {24.35}  & {22.25}  & {17.60}  & {22.93}  & {25.52}  & {20.62}  & {22.76}  & {20.39}  & \cellcolor{gris2}{22.14} \\
\hline
\hline
\end{tabular}}
\vspace{0.01cm}
\caption{\textbf{Qualitative and quantitative evaluation on image in-painting.} \textbf{Top:} From left to right, we display ground truth, input and result for the Barbara image for a 50\% missing pixels. \textbf{Bottom:} The table reports the PSNR in dB (higher is better) using our approach for 10 images. We indicate the solution for a missing pixel rate of $\{30\%,50\%,60\%\}$.}
\label{table_inpainting}
\end{table}

\section{Conclusion}
\label{sec:conclusion}

LRD is a powerful framework that provides a sufficient prior to learn the latent structure of data in multidimensional settings. The results obtained regarding the compressed video reconstruction verify our claims. Moreover its formulation is flexible enough to deal with incomplete data allowing its application in tensor completion problems as we verified with the experiments regarding image in-painting. As a future work would be interesting to evaluate how this approach deals with increasing data dimensions and large datasets, as in this situations data compression is of an important matter.



\bibliographystyle{IEEEbib}
\bibliography{refs}

\end{document}

%% file: newcommands2.tex
\newcommand{\comment}[1]{}

\newcommand{\plusc}{\hspace{-0.5mm}+\hspace{-0.5mm}}

\newcommand{\bs}{\mathbf{s}}

\newcommand{\bu}{\mathbf{u}}

\newcommand{\bv}{\mathbf{v}}

\newcommand{\bx}{\mathbf{x}}

\newcommand{\bz}{\mathbf{z}}

\newcommand{\bA}{\mathbf{A}}

\newcommand{\bD}{\mathbf{D}}

\newcommand{\bF}{\mathbf{F}}

\newcommand{\bI}{\mathbf{I}}
\newcommand{\bJ}{\mathbf{J}}
\newcommand{\bK}{\mathbf{K}}

\newcommand{\bP}{\mathbf{P}}
\newcommand{\bQ}{\mathbf{Q}}

\newcommand{\bS}{\mathbf{S}}
\newcommand{\bT}{\mathbf{T}}

\newcommand{\bU}{\mathbf{U}}

\newcommand{\bW}{\mathbf{W}}
\newcommand{\bX}{\mathbf{X}}
\newcommand{\bY}{\mathbf{Y}}

\newcommand{\argmin}{\operatornamewithlimits{arg\,min}}

%% file: newcommands3.tex
\makeatletter
\DeclareRobustCommand\onedot{\futurelet\@let@token\@onedot}
\def\@onedot{\ifx\@let@token.\else.\null\fi\xspace}

\def\ie{\emph{i.e}\onedot}

\makeatother
